\documentclass[11pt]{article}

\usepackage[preprint]{acl}

\usepackage{times}
\usepackage{latexsym}

\usepackage{amssymb} 
\usepackage{colortbl} 

\usepackage[T1]{fontenc}
\usepackage[utf8]{inputenc}

\usepackage{microtype}

\usepackage{inconsolata}

\usepackage{graphicx}

\usepackage{pifont}

\usepackage{booktabs}
\newcommand{\gpticon}{\smash{\raisebox{-0.2ex}{\includegraphics[height=1.2em]{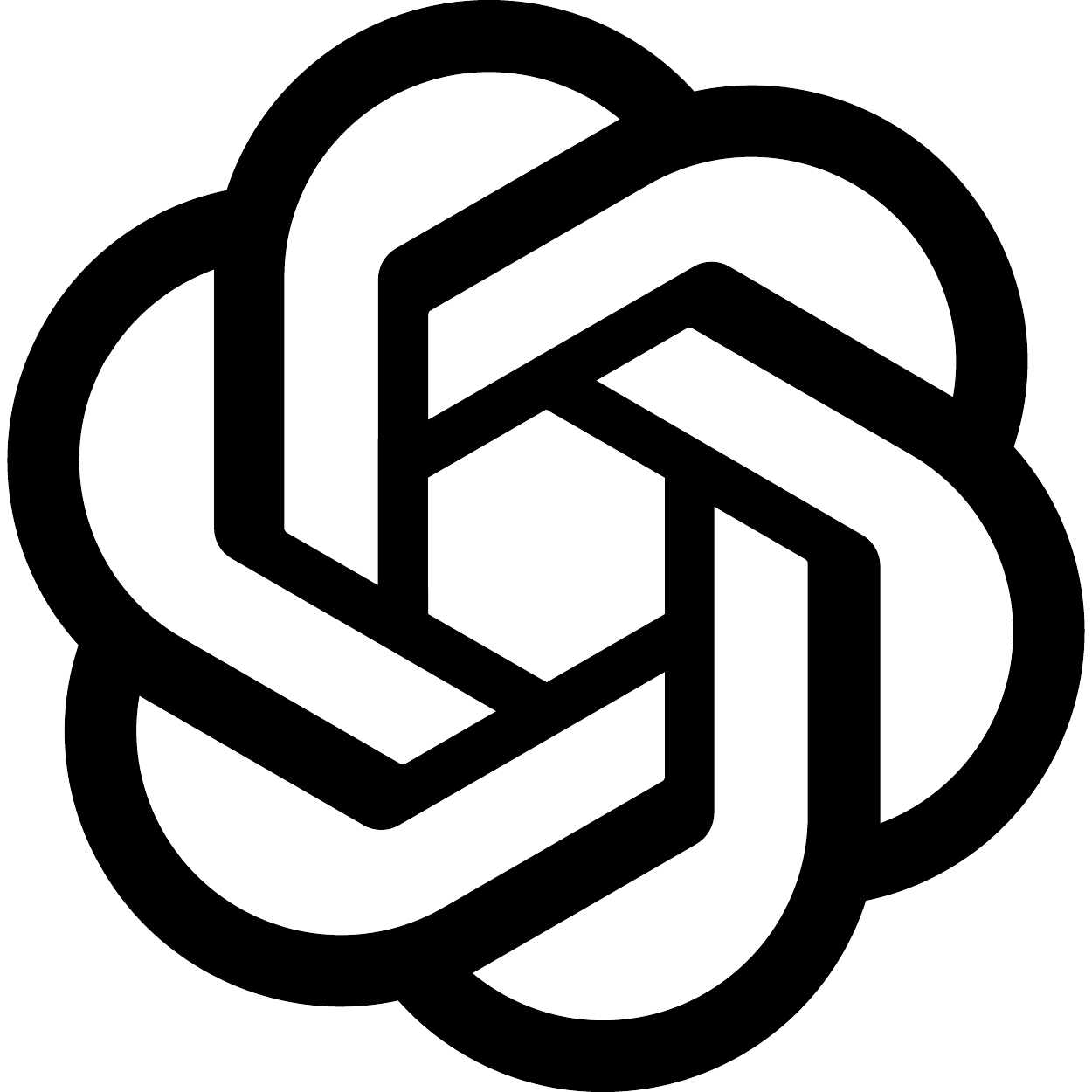}}}}
\newcommand{\geminiicon}{\smash{\raisebox{-0.2ex}{\includegraphics[height=1.2em]{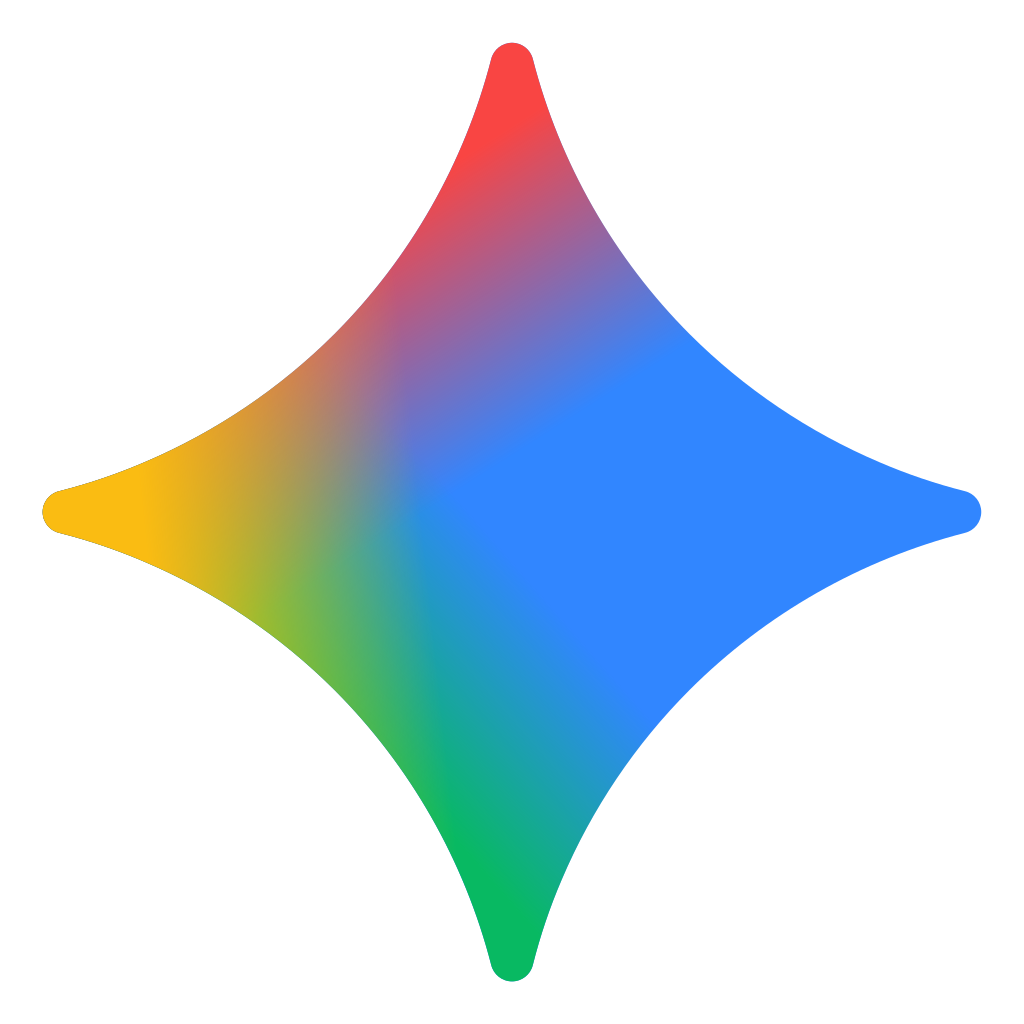}}}}
\newcommand{\llamaicon}{\smash{\raisebox{-0.2ex}{\includegraphics[height=1.2em]{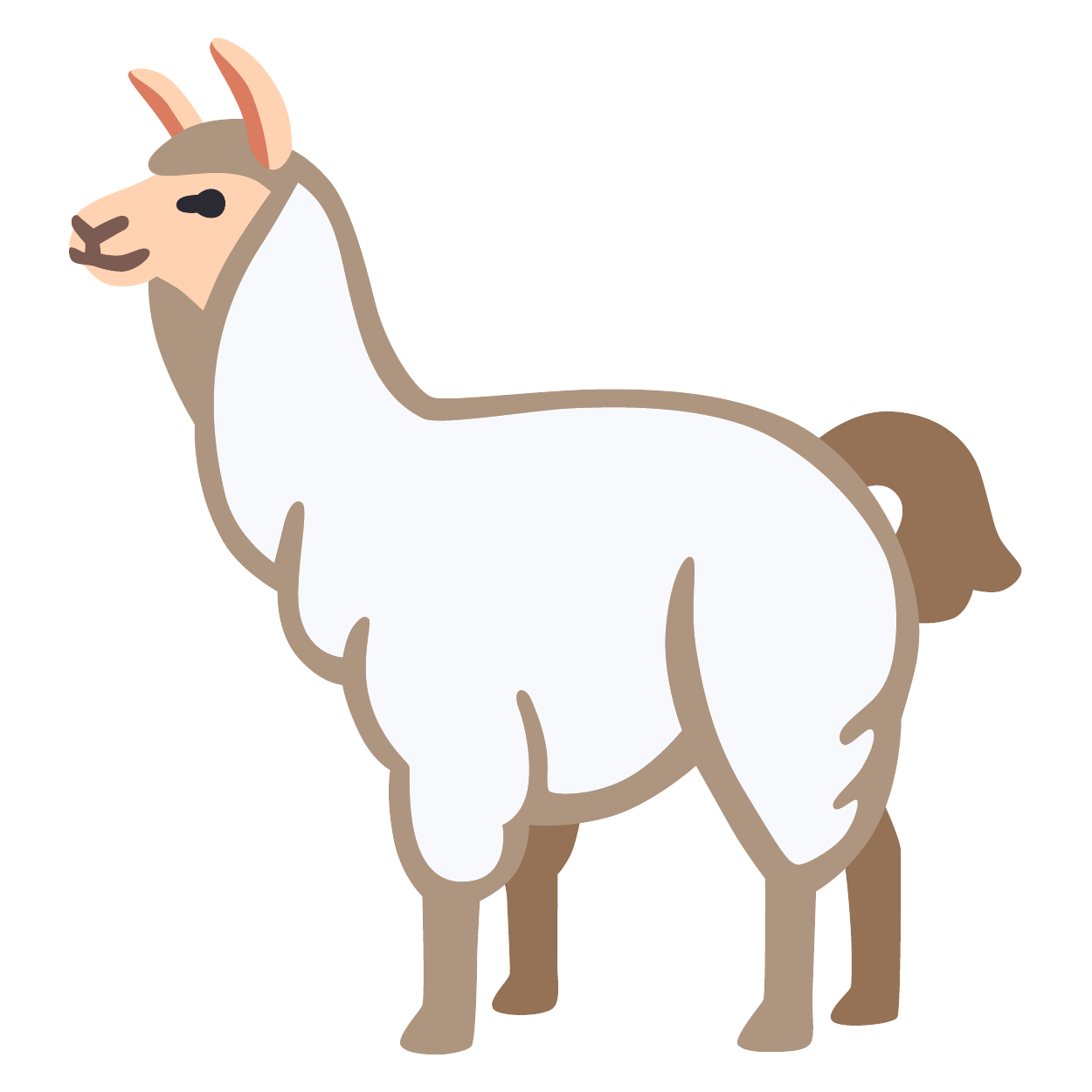}}}}
\newcommand{\googleicon}{\smash{\raisebox{-0.2ex}{\includegraphics[height=1.2em]{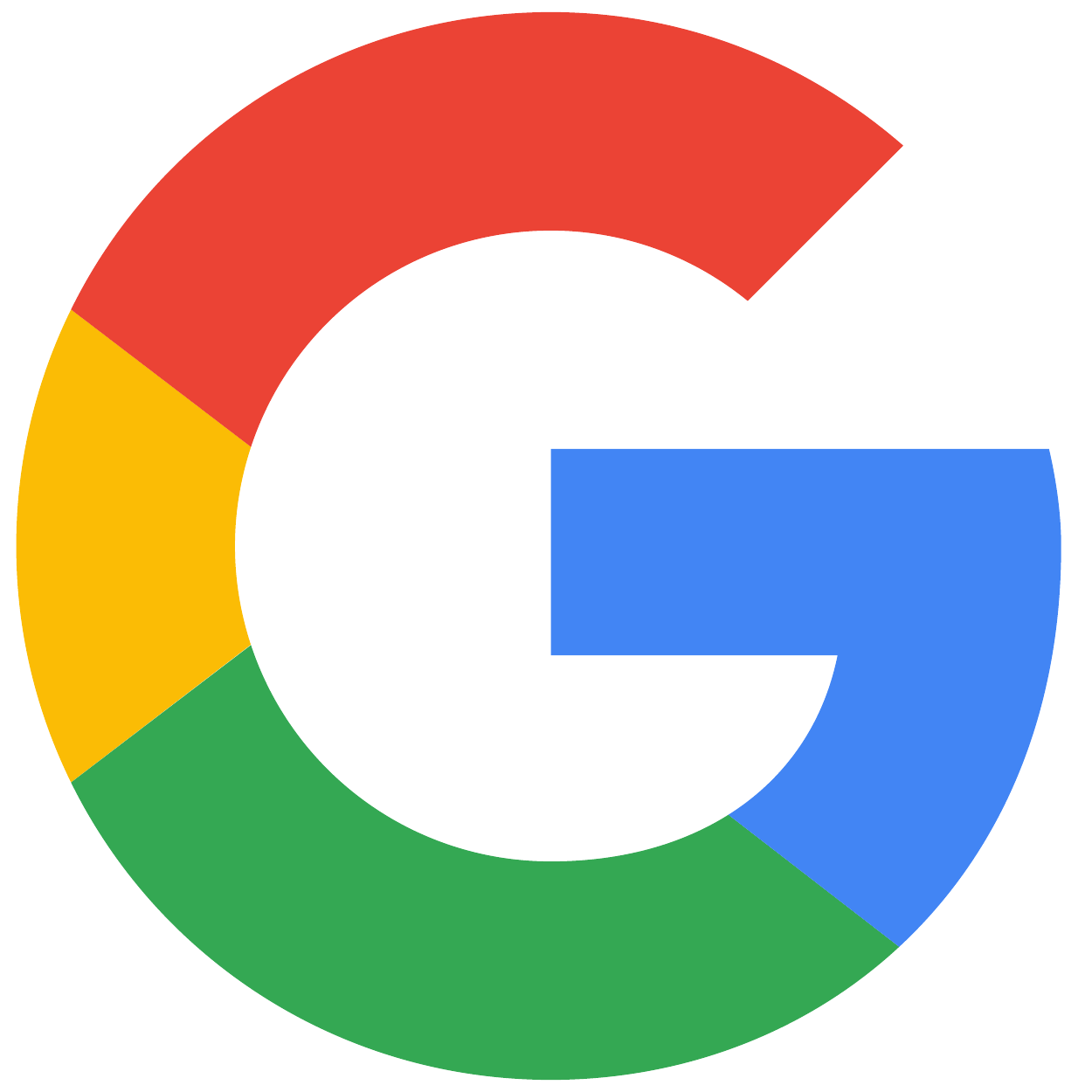}}}}
\newcommand{\metaicon}{\smash{\raisebox{-0.2ex}{\includegraphics[height=1.2em]{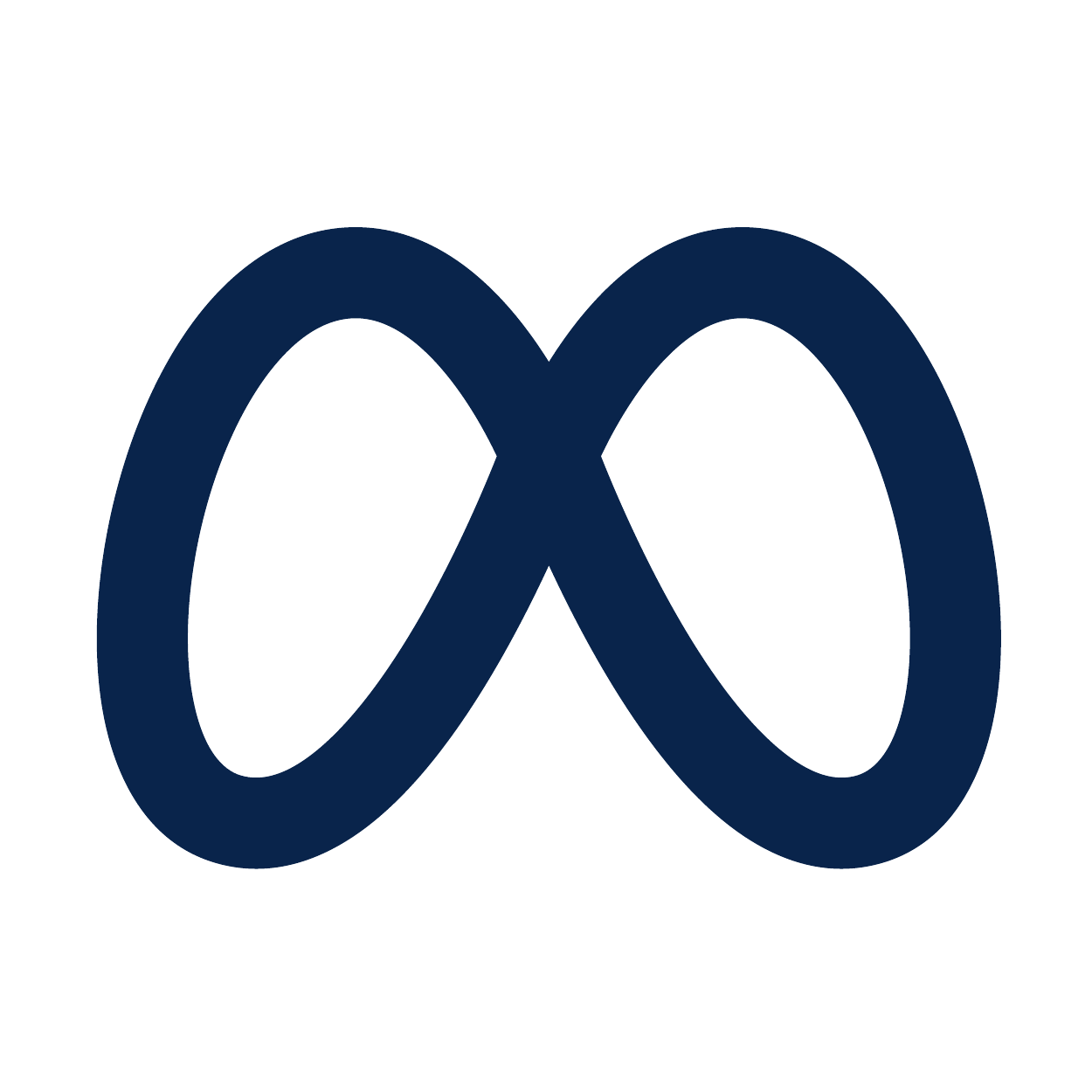}}}}
\newcommand{\openaiicon}{\smash{\raisebox{-0.2ex}{\includegraphics[height=1.2em]{assets/openai.pdf}}}}

\usepackage[most]{tcolorbox}

\title{AiMNLP at BHASHA Task 1 IndicGEC: ``When Data is Scarce, Prompt Smarter''... Approaches to Grammatical Error Correction in Low-Resource Settings}

\author{
  \textbf{Somsubhra De\textsuperscript{1}},
  \textbf{Harsh Kumar\textsuperscript{1}}\and
  \textbf{Arun Prakash A\textsuperscript{1,2}}
\\
\\
  \textsuperscript{1}IIT Madras
  \textsuperscript{2}AI4Bharat
}

\begin{document}
\maketitle
\begin{abstract}
Grammatical error correction (GEC) is an important task in Natural Language Processing that aims to automatically detect and correct grammatical mistakes in text. While recent advances in transformer-based models and large annotated datasets have greatly improved GEC performance for high-resource languages such as English, the progress has not extended equally. For most Indic languages, GEC remains a challenging task due to limited resources, linguistic diversity and complex morphology. In this work, we explore prompting-based approaches using state-of-the-art large language models (LLMs), such as GPT-4.1, Gemini-2.5 and LLaMA-4, combined with few-shot strategy to adapt them to low-resource settings. We observe that even basic prompting strategies, such as zero-shot and few-shot approaches, enable these LLMs to substantially outperform fine-tuned Indic-language models like Sarvam-22B, thereby illustrating the exceptional multilingual generalization capabilities of contemporary LLMs for GEC. Our experiments show that carefully designed prompts and lightweight adaptation significantly enhance correction quality across multiple Indic languages. We achieved leading results in the shared task--ranking \textit{1st in Tamil (GLEU: 91.57) and Hindi (GLEU: 85.69), 2nd in Telugu (GLEU: 85.22), 4th in Bangla (GLEU: 92.86), and 5th in Malayalam (GLEU: 92.97)}. These findings highlight the effectiveness of prompt-driven NLP techniques and underscore the potential of large-scale LLMs to bridge resource gaps in multilingual GEC.
\end{abstract}

\section{Introduction}
Grammar forms the foundation of any language-- it plays a fundamental role in ensuring clarity, fluency and precision in written communication. In Natural Language Processing (NLP), improving the grammatical quality of written text has become an important research focus, leading to the development of Grammatical Error Correction (GEC) systems. At the sentence level, a GEC system takes a potentially erroneous sentence as input and aims to generate a corrected version that is fluent, accurate and semantically faithful to the original. In practice, this involves detecting and correcting different kinds of errors such as spelling mistakes, punctuation issues, grammatical inconsistencies, issues related to verb usage (correct conjugation, tense, aspect, agreement with the subject), gender and pronoun agreement errors, as well as inappropriate word choices. The scope of GEC extends beyond simple grammar checking - it encompasses the broader goal of making a sentence more natural and comprehensible while preserving its intended meaning.

Despite its apparent simplicity, defining what constitutes `grammatical correctness' is not straightforward! Grammaticality can be subjective and context-dependent, especially in languages with diverse dialects, flexible word order or rich morphology. For Indic languages such as Tamil, Malayalam, Hindi, Bangla, and Telugu, that we work on, this complexity is even more pronounced. These languages exhibit significant variation in syntax, inflection and script, which makes the task of automatic correction far more challenging than in English or other high-resource languages. Furthermore, the scarcity of large annotated corpora, standardized benchmarks and robust evaluation metrics adds to the difficulty of developing effective GEC systems for Indic languages.

In recent years, transformer-based models and large language models (LLMs) have shown impressive results for sentence-level GEC in English and other major languages \cite{tarnavskyi-etal-2022-ensembling,rothe-etal-2021-simple}. However, extending these advances to Indic languages requires adapting such models to low-resource settings and accounting for linguistic diversity. We explore prompting-based approaches, designing carefully crafted instructions to guide LLMs (GPT-4 and Gemini) towards effective correction. This work explores how prompt-driven and lightweight fine-tuning techniques can enhance sentence-level GEC for multiple Indic languages. We aim to show that, with thoughtful design and adaptation, modern LLMs can effectively handle the unique challenges of multilingual error correction and help bridge the resource gap between high and low-resource languages.\footnote{Codes and data can be found at \url{https://github.com/somsubhra04/BHASHA2025}}
\section{Related work}
Early research on Machine Translation (MT) laid much of the groundwork for sentence-level transformation tasks like GEC. Traditional rule-based and statistical MT models focused on translating between languages by learning alignment patterns between source and target sentences. These approaches, though effective in specific domains, were often limited by their reliance on handcrafted rules and parallel corpora.

The advent of Neural Machine Translation (NMT) fundamentally changed the landscape of text generation and correction tasks. Sequence-to-sequence architectures with attention mechanisms \cite{bahdanau2016neuralmachinetranslationjointly,luong-etal-2015-effective} enabled models to learn contextual dependencies more effectively, producing fluent and coherent outputs.

Subsequent developments introduced Transformer-based models \cite{vaswani2023attentionneed}, which achieved significant performance improvements across both MT and GEC. Pre-trained language models such as BERT \cite{devlin2019bertpretrainingdeepbidirectional}, RoBERTa \cite{liu2019robertarobustlyoptimizedbert} and T5 \cite{raffel2023exploringlimitstransferlearning} further advanced GEC research by providing strong contextual representations. These models have been fine-tuned for grammatical correction tasks, achieving promising results on benchmarks like CoNLL-2014 and BEA-2019.

Parallel to model innovations, researchers also focused on data augmentation and noise generation strategies to address the scarcity of large-scale annotated corpora. Synthetic error generation \cite{awasthi-etal-2019-parallel} --by introducing artificial grammatical mistakes into clean corpora-proved effective for pre-training and domain adaptation, particularly in low-resource settings. Such approaches allowed models to generalize better to real-world learner errors while mitigating overfitting to limited gold-standard data.

More recently, multilingual and low-resource GEC have gained attention, driven by the growing need to support diverse languages beyond English. Approaches leveraging multilingual transformers such as mBERT and XLM-R \cite{conneau2020unsupervisedcrosslingualrepresentationlearning} have shown promise for languages with limited annotated data.
\section{Dataset}
\begin{figure*}[ht]
  \includegraphics[width=1.0\linewidth]{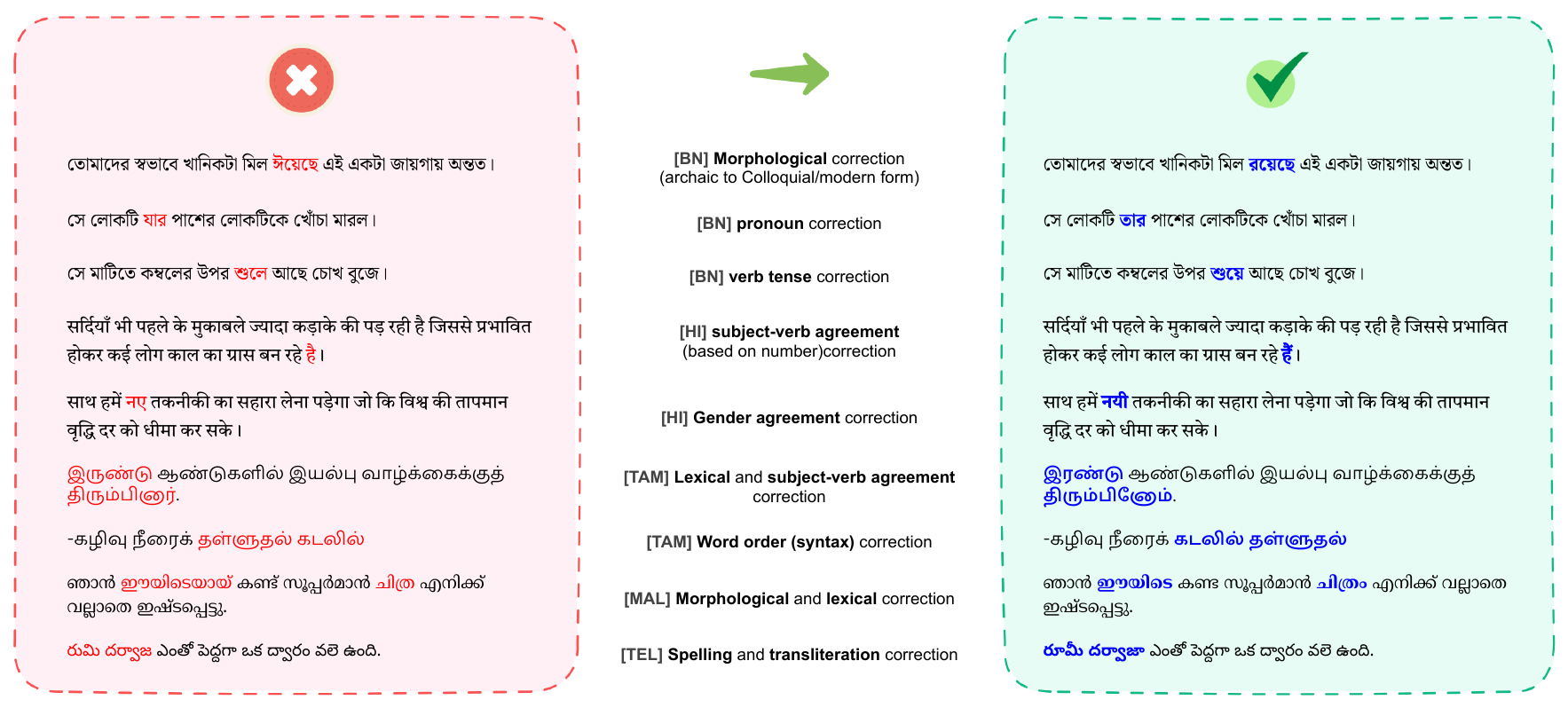}
  \caption {\label{egs} Examples from the GEC task dataset. Input sentence \textcolor{red}{\ding{55}} (with errors in \textcolor{red}{red}) - ground truth \textcolor{green}{\checkmark} (with corrections in \textcolor{blue}{blue}) pairs. Error types have been mentioned, based on our understanding.}
\end{figure*}

We utilize the multilingual datasets from the \href{https://bhasha-workshop.github.io/}{Workshop on Benchmarks, Harmonization, Annotation and Standardization for Human-Centric AI in Indian Languages (BHASHA)}. The dataset covers five major Indian languages - Tamil \textsc{(Tam)}, Malayalam \textsc{(Mal)}, Hindi \textsc{(Hi)}, Bengali \textsc{(Bn)} and Telugu \textsc{(Tel)}. Each language dataset consists of parallel sentence pairs designed for the GEC task. Each instance contains two columns: \textcolor{red}{\textit{Input sentence}}: a sentence (erroneous) containing one or more grammatical, spelling or lexical errors and
\textcolor{green}{\textit{Output sentence}}: the corresponding grammatically correct version produced by human annotators. (refer to Fig~\ref{egs} for examples) The datasets were created by asking native speakers of the respective languages to write essays on given topics in an exam-like setting, after which language experts manually corrected the errors to produce the gold-standard outputs. The overall data distribution across languages is presented in Table~\ref{tab:dataset_stats}.
\begin{table}[ht]
\centering
\begin{tabular}{lccccc}
\hline
\rowcolor{blue!10}
\textbf{} & \textsc{Tam} & \textsc{Mal} & \textsc{Hi} & \textsc{Bn} & \textsc{Tel} \\ 
\hline
Train & 91 & 300 & 599 & 598 & 599 \\
Dev   & 16 & 50 & 107 & 101 & 100 \\
\textbf{Test}  & \textbf{65} & \textbf{102} & \textbf{236} & \textbf{330} & \textbf{310} \\
\hline
Total  & 172 & 452 & 942 & 1029 & 1009 \\
\hline
\end{tabular}
\caption{Dataset statistics for Indic-GEC tasks. Each sample is a pair of erroneous and corrected sentences.}
\label{tab:dataset_stats}
\end{table}

\section{Methodology}

The Indic-GEC task is formulated as a sequence-to-sequence transformation problem, where a potentially erroneous sentence 
\( x = (x_1, x_2, \ldots, x_n) \) in an Indic language is mapped to its grammatically corrected version 
\( y = (y_1, y_2, \ldots, y_m) \). 
The objective of the model is to estimate the conditional probability distribution 
\( P_\theta(y|x) \), parameterized by \(\theta\), that maximizes the likelihood of the correct output sequence:
\begin{equation}
\hat{y} = \arg\max_{y} P_\theta(y|x)
\end{equation}
\subsection{Problem Formulation}
Given a parallel corpus 
\( D = \{(x^{(i)}, y^{(i)})\}_{i=1}^N \),
where \(x^{(i)}\) is a noisy input sentence and \(y^{(i)}\) is its grammatically corrected counterpart,
the model is trained to minimize the negative log-likelihood (NLL) of the target sequence conditioned on the input:
\begin{equation}
\mathcal{L}_{\text{GEC}}(\theta) = -\sum_{i=1}^{N} \sum_{t=1}^{|y^{(i)}|} 
\log P_\theta\left(y_t^{(i)} \mid y_{<t}^{(i)}, x^{(i)} \right)
\end{equation}
This formulation enables the model to capture token-level dependencies that reflect both syntactic and morphological coherence in Indic scripts.
\subsection{LLM-based Inference Framework}
We utilize three large language models -- \gpticon{} \href{https://platform.openai.com/docs/models/gpt-4.1-mini}{\textbf{GPT-4.1 mini}}, \geminiicon{} \textbf{Gemini-2.5-Flash} \cite{comanici2025gemini25pushingfrontier} and \llamaicon{} \href{https://ai.meta.com/blog/llama-4-multimodal-intelligence/}{\textbf{Llama-4-Maverick-17B-128E-Instruct}} for inference-only Indic-GEC in \textit{zero-shot} and \textit{few-shot} prompting paradigms. 
These models are employed as instruction-following LLMs with role-based prompts (check Appendix ~\ref{sec:appendix}) to evaluate their grammatical correction capabilities across multiple Indic languages.

\begin{table}[t]
\centering
\small

\begin{tabular}{l l}
\hline
\rowcolor{black!5}
\textbf{Parameter} & \textbf{Value / Setting} \\
\hline
Base model & \texttt{sarvamai/sarvam-m} \\
Dataset & Hi-GEC \\
Epochs & 25 \\
Batch size (per device) & 1 \\
Gradient accumulation & 16 steps \\
Effective batch size & 16 \\
Learning rate & \(2\times10^{-4}\) \\
Optimizer & Paged AdamW (32-bit) \\
Scheduler & Cosine with warmup ratio = 0.03 \\
Quantization & 4-bit NF4 (QLoRA) \\
Compute dtype & bfloat16  \\
LoRA rank (\(r\)) & 64 \\
LoRA $\alpha$ & \(2r = 128\) \\
LoRA dropout & 0.05 \\
Bias & none \\
GLEU Score & 38.56 \\
\hline
\end{tabular}

\caption{Hyperparameter configuration for fine-tuning the \textbf{Sarvam-M 24B} model on the Hi-GEC dataset using LoRA.}
\label{tab:sarvam_hyperparams}
\end{table}\

\subsubsection{Zero-Shot Prompting}

In the zero-shot setting, the model receives only the input sentence along with an instruction prompt:

{\small
\begin{equation}
M(x) = \text{LLM}\big([\text{System Prompt: GEC Instruction}], x\big)
\end{equation}
}
where \(M(x)\) denotes the model output for the input \(x\). 
The prompt explicitly guides the model to correct grammatical, morphological, spacing, and diacritic errors while retaining the meaning, tone and script of the original text.

\begin{table*}[ht]
  \centering
  \small
  \setlength{\tabcolsep}{4pt}
  \begin{tabular}{l|ccc|ccc|ccc}
    \hline
    \textbf{Model}
    & \multicolumn{3}{c|}{\textbf{\textsc{Tam}}}
    & \multicolumn{3}{c|}{\textbf{\textsc{Mal}}}
    & \multicolumn{3}{c}{\textbf{\textsc{Hi}}} \\
    \cline{2-10}
    & \textbf{GLEU} & \textbf{$F_{0.5}$} & \textbf{BERT-score}
    & \textbf{GLEU} & \textbf{$F_{0.5}$} & \textbf{BERT-score}
    & \textbf{GLEU} & \textbf{$F_{0.5}$} & \textbf{BERT-score} \\
    \hline
    \googleicon Gemini-2.5-Flash \textit{(fs)} & \cellcolor{green!30}\textbf{\textcolor{blue}{\underline{91.57}}} & 87.82& 97.83 & \cellcolor{green!30}\textbf{92.97} & 88.48& 97.89 & 84.61 & 88.01& 95.69\\
    \openaiicon GPT-4.1 mini \textit{(fs)} & 86.00 & 78.97& 96.52 & 91.78 &  84.72& 97.08 & \cellcolor{green!30}\textbf{\textcolor{blue}{\underline{85.69}}} & 87.86& 95.76\\
    \openaiicon GPT-4.1 mini \textit{(zs)} & 85.51 & 78.45& 96.38 & 92.34 & 84.62& 97.24 & 85.37 &  87.80& 95.53\\
    \metaicon LLaMA-4 maverick \textit{(zs)} &  88.70& 81.50&  96.84&  92.68& 85.38&  97.20&  83.10& 86.04& 94.64\\
    \metaicon LLaMA-4 maverick \textit{(fs)} &  85.62& 77.75& 95.98&  90.75& 83.22&  96.65&  85.37&  87.35& 95.56\\
    \hline
  \end{tabular}

  \vspace{0.5em}

  \begin{tabular}{l|ccc|ccc}
    \hline
    \textbf{Model}
    & \multicolumn{3}{c|}{\textbf{\textsc{Bn}}}
    & \multicolumn{3}{c}{\textbf{\textsc{Tel}}} \\
    \cline{2-7}
    & \textbf{GLEU} & \textbf{$F_{0.5}$} & \textbf{BERT-score}
    & \textbf{GLEU} & \textbf{$F_{0.5}$} & \textbf{BERT-score} \\
    \hline
    \googleicon Gemini-2.5-Flash \textit{(fs)} & 92.23 & 89.61& 97.10 & 84.16 & 76.96& 94.92\\
    \openaiicon GPT-4.1 mini \textit{(fs)} & \cellcolor{green!30}\textbf{92.86} & 89.27& 97.35 & \cellcolor{green!30}\textbf{85.22} & 77.68& 95.28\\
    \openaiicon GPT-4.1 mini \textit{(zs)} & 91.62 & 86.98& 96.79 & 84.74 & 76.75& 95.15\\
    \metaicon LLaMA-4 maverick \textit{(zs)} &  90.39& 86.48&  96.30&  83.01& 74.20&  94.28\\
    \metaicon LLaMA-4 maverick \textit{(fs)} &  92.00& 88.02& 97.19&  82.02& 74.28&  94.08\\
    \hline
  \end{tabular}

  \caption{\label{tab:test_results}
    Performance of different approaches on the \textbf{test set} across languages.
    Top table shows results for Tamil, Malayalam, and Hindi; bottom table for Bengali and Telugu.
    \colorbox{green!30}{Highlighted cells} indicate the best-performing model for each language, while \colorbox{green!30}{\textcolor{blue}{\underline{underlined values}}} denote the overall best score in the task.
  }
\end{table*}

\subsubsection{Few-Shot Prompting}

In few-shot prompting, the model is provided with \(k\) in-context examples, enabling implicit adaptation to the Indic grammatical structure:

{\small
\begin{equation}
M_k(x) = \text{LLM}\big([\text{System Prompt}], \{(x_1, y_1), \ldots, (x_k, y_k)\}, x\big)
\end{equation}
}
Each example pair \((x_i, y_i)\) represents a noisy and corrected sentence. For Hindi, 10 such examples from the train set were manually curated to cover orthographic, matra and spacing errors, while for the other languages 10 examples were drawn randomly from their respective training sets. This allowed the model to infer task-specific patterns and linguistic consistency directly from context.
\subsection{Fine-Tuned Sarvam-M Model}
In addition to zero and few-shot LLM-based inference, we \href{https://huggingface.co/Harsh0304/Sarvam-Hi-gec-finetuned}{fine-tuned} the \textbf{Sarvam-M 24B}\footnote{The model card can be found at \url{https://huggingface.co/sarvamai/sarvam-m}} multilingual model using the \textbf{Hi-GEC} dataset 
\citep{sharma-bhattacharyya-2025-hi}, which contains parallel Hindi sentences in \texttt{.src} and \texttt{.tgt} format 
for training, validation, and testing.

Given the large parameter size, we adopt the \textbf{Low-Rank Adaptation (LoRA)} \cite{hu2021loralowrankadaptationlarge} technique 
for parameter-efficient fine-tuning. 
Let \(W_0 \in \mathbb{R}^{d \times k}\) denote the frozen pretrained weight matrix of a \textit{decoder block}. 
LoRA introduces a low-rank decomposition to approximate the weight update:
\begin{equation}
\Delta W = B A
\end{equation}
where \(A \in \mathbb{R}^{r \times k}\), \(B \in \mathbb{R}^{d \times r}\), and \(r \ll \min(d, k)\). 
The adapted weight becomes:
\begin{equation}
W = W_0 + \alpha \cdot \Delta W
\end{equation}
Here, \(\alpha\) controls the contribution of the LoRA adapter to the overall weight update. 
In our setup, a rank of \(r = 64\) is used, and the model is fine-tuned for 25 epochs using the same cross-entropy loss defined in Equation~(2). 
Only the low-rank matrices \(A\) and \(B\) are trainable, while all base model parameters remain frozen, significantly reducing memory consumption and training cost.

The fine-tuned Sarvam-M model serves as a domain-adapted reference for Hindi grammatical correction. 
The detailed hyperparameter configuration (learning rate, adapter placement, and optimizer setup) 
is described separately in Table~\ref{tab:sarvam_hyperparams}.

\section{Results \& Analysis}
\subsection{Evaluation Metrics}
We evaluate the model performance using a combination of metrics: GLEU (1–4 gram) for overall fluency and faithfulness, {$F_{0.5}$} (useful for precision-focused correction accuracy) \& BERT-Score (F1) for the semantic similarity between outputs \& references. The scores for the submitted runs, on the test set are mentioned in Table~\ref{tab:test_results}. We observe that Gemini-2.5-Flash consistently achieves the highest GLEU for Tamil (\textbf{91.57}) and Malayalam (\textbf{92.97}), indicating strong alignment with reference corrections in these low-resource Dravidian languages. Notably, LLaMA-4 maverick proves highly effective in this cluster, outperforming GPT-4.1 mini in both Tamil and Malayalam to secure the second-highest scores. We observe that LLaMA achieves superior generalization in zero-shot settings for Dravidian languages, likely because the few-shot exemplars introduce stylistic noise that disrupts the model's otherwise robust internal multilingual representations. For Indo-Aryan languages, the trend shifts... in Hindi, Gemini-2.5-Flash slightly lags behind GPT-4.1 mini-- which attains the highest GLEU (\textbf{85.69}). In contrast, the finetuned Sarvam-M lags significantly and fails to capture the correct edits, achieving only \textbf{{\textcolor{red}{13.81}}} GLEU in Hindi--we attribute this to its limited total parameter capacity compared to the massive sparse architectures of the foundation models, which restricts its ability to resolve complex, long-range grammatical dependencies. For Bengali and Telugu, GPT-4.1 mini slightly outperforms Gemini in GLEU.

Tokenization plays a crucial role in evaluating GEC models for Indic languages. Unlike English and many European languages--where standard tokenizers (e.g., those in NLTK) perform well—Indic languages pose additional challenges due to complex scripts \& rich morphology. A simple or naïve tokenization approach may incorrectly split or merge tokens, leading to unreliable n-gram matching and inflated or deflated GLEU scores. For more robust handling, IndicNLP tokenizer \cite{kakwani2020indicnlpsuite} is a better option that supports Indic languages. This ensures fairer evaluation across languages and improves the reliability of metric-based comparisons.

The results demonstrate that carefully designed prompts, including multiple styles and refined instructions, substantially improve GEC performance across languages, with zero-shot performance consistently lower, highlighting the benefit of providing context through few-shot prompting.
\subsection{Qualitative analysis}
We analyzed where the LLMs tended to fail and noticed some recurring patterns. In several cases, the reference and input sentences were identical, which prompted us to update the prompt instructions to explicitly return the text un-changed when no corrections were needed. (Check Fig~\ref{A-2}) For Hindi, the models occasionally confused orthographic variants that sound similar but differ in spelling conventions and frequently omitted diacritic marks, particularly in transliterated or mixed-script tokens. They sometimes missed plural or oblique forms, tending to normalize words rather than applying the correct inflection. While the models generally preserved the lexical content, they occasionally failed to restructure clauses syntactically, revealing limitations in hierarchical parsing. Minor issues such as unnecessary quotation marks, missing spaces around punctuation, or preference for modern spellings over strict orthography were also observed. These patterns suggest that, despite capturing many grammatical structures, LLMs can falter on script consistency, inflectional morphology, and fine-grained syntax.
\section{Conclusion and Future Work}

This study presented a comprehensive framework for Indic Grammatical Error Correction (Indic-GEC) that integrates both inference-based large language models and parameter-efficient fine-tuning of a large-scale multilingual model. Through zero-shot and few-shot prompting, we explored the intrinsic grammatical reasoning abilities of some of the SoTA instruction-tuned LLMs across five major Indic languages. Complementarily, the Sarvam-M 24B model was fine-tuned using a LoRA-based adaptation strategy on the Hi-GEC corpus, enabling domain-specific correction for Hindi while preserving computational efficiency.

The proposed approach highlights the strengths and limitations of contemporary multilingual LLMs in addressing the unique orthographic, morphological, and syntactic characteristics of Indic languages. The results emphasize that while general-purpose models demonstrate strong grammatical awareness in high-resource settings, language-specific fine-tuning remains essential for achieving robust and culturally consistent error correction in low-resource Indic contexts.

In future work, we aim to extend this framework towards a comprehensive \textbf{cross-lingual analysis} of grammatical generalization across Indo-Aryan and Dravidian language families. This will involve evaluating the transferability of learnt correction patterns from Hindi to other related languages such as Marathi and Bengali, and from Tamil or Telugu to low-resource southern languages. Additionally, we plan to explore \textbf{multilingual joint fine-tuning} strategies using shared sub-word representations to enhance cross-lingual transfer and minimize model bias.

A critical dimension of this exploration will be a rigorous investigation into tokenizer fertility and its correlation with downstream performance. While recent studies \cite{ali-etal-2024-tokenizer} suggest that high fertility scores can disproportionately degrade model performance on generation tasks for morphologically rich languages, we did not conduct a systematic ablation of this relationship here. Optimizing tokenizer vocabularies specifically for Dravidian agglutination could yield significant gains in both inference efficiency and correction quality. Finally, incorporating syntactic and morphological constraints during decoding, along with leveraging bilingual and transliterated corpora, remains a key avenue to further improve the grammatical fluency and fidelity of corrections.

These findings indicate that instruction-tuned LLMs currently offer more reliable grammatical correction capabilities than resource-constrained fine-tuning of very large multilingual models.

\section*{Limitations}
The experiments were constrained by compute availability, API cost and call rates, which limited the number of runs, model variants and prompt iterations tested across languages. Prompt sensitivity occasionally introduced variability across runs. Since the authors were not uniformly proficient across all languages, qualitative analysis -- including a deeper examination of error types and correction patterns could not be done for each language. Furthermore, larger and more diverse multilingual GEC benchmarks would have enabled stronger conclusions and fairer cross-lingual comparison.

\section*{Acknowledgments}
We thank the BHASHA organizers for their for their prompt correspondence and the anonymous reviewers, shared-task PCs for their thoughtful feedback.
\nocite{bhattacharyya-bhattacharya-2025-leveraging}
\bibliography{custom}

\appendix
\section{Prompts}
\label{sec:appendix}
The best-performing system prompts from our experiments are provided below (check \ref{tab:prompt_egs}).
\begin{table*}[ht]
\captionsetup{labelformat=empty}
\caption{}\label{tab:prompt_egs}
\centering
\begin{tcolorbox}[
    colback=lightgray!20!white,
    colframe=black,
    title=\textbf{TAM \& MAL (Gemini, Few-shot)},
    fonttitle=\small\bfseries,
    boxsep=5pt,
    arc=1pt,
    leftrule=1pt, rightrule=1pt, toprule=1pt, bottomrule=1pt
]
\ttfamily
``````You are a <Language> Grammatical Error Correction assistant, in low resource settings. Your task is to accurately identify and correct grammatical errors in the given <Language> sentence. Correct all types of grammatical errors:\\
    \textbf{Verb usage}: Correct conjugation, tense, aspect, and agreement with the subject.,\\
    \textbf{Pronouns}: Usage of proper personal, possessive, and reflexive pronouns.,\\
    \textbf{Prepositions}: Correct use of postpositions or prepositions in context.,\\
    Fix spelling mistakes, diacritic marks (matras), and punctuation errors.,\\
    \textbf{Gender and number agreement}: Ensure adjectives, nouns, and verbs match in gender (masculine/feminine) and number (singular/plural).,\\
    The output should be ONLY the CORRECTED sentence, without any extra text or explanation. \textit{If the input is already correct, return it unchanged.} Please ensure the corrections follow the rules and preserve the intended meaning.\\ Below are 10 random sentences for your reference.''''''
\end{tcolorbox}

\begin{tcolorbox}[
    colback=lightgray!20!white,
    colframe=black,
    title=\textbf{HI, BN \& TEL (GPT, Few-shot)},
    fonttitle=\small\bfseries,
    boxsep=5pt,
    arc=1pt,
    leftrule=1pt, rightrule=1pt, toprule=1pt, bottomrule=1pt
]
\ttfamily
``````You are a Grammatical Error Correction (GEC) assistant for low-resource Indian languages.\\
Your job: correct only grammar, spelling, spacing, matras/diacritics, punctuation, and light word-form errors.\\
Do NOT translate. Preserve the meaning, script, and style of the input language.\\
Return ONLY the corrected sentence with no quotes, no labels, no extra text.\\
If the input is already correct, return it unchanged.''''''
\end{tcolorbox}
\end{table*}

\begin{figure*}[ht]
  \includegraphics[width=1.0\linewidth]{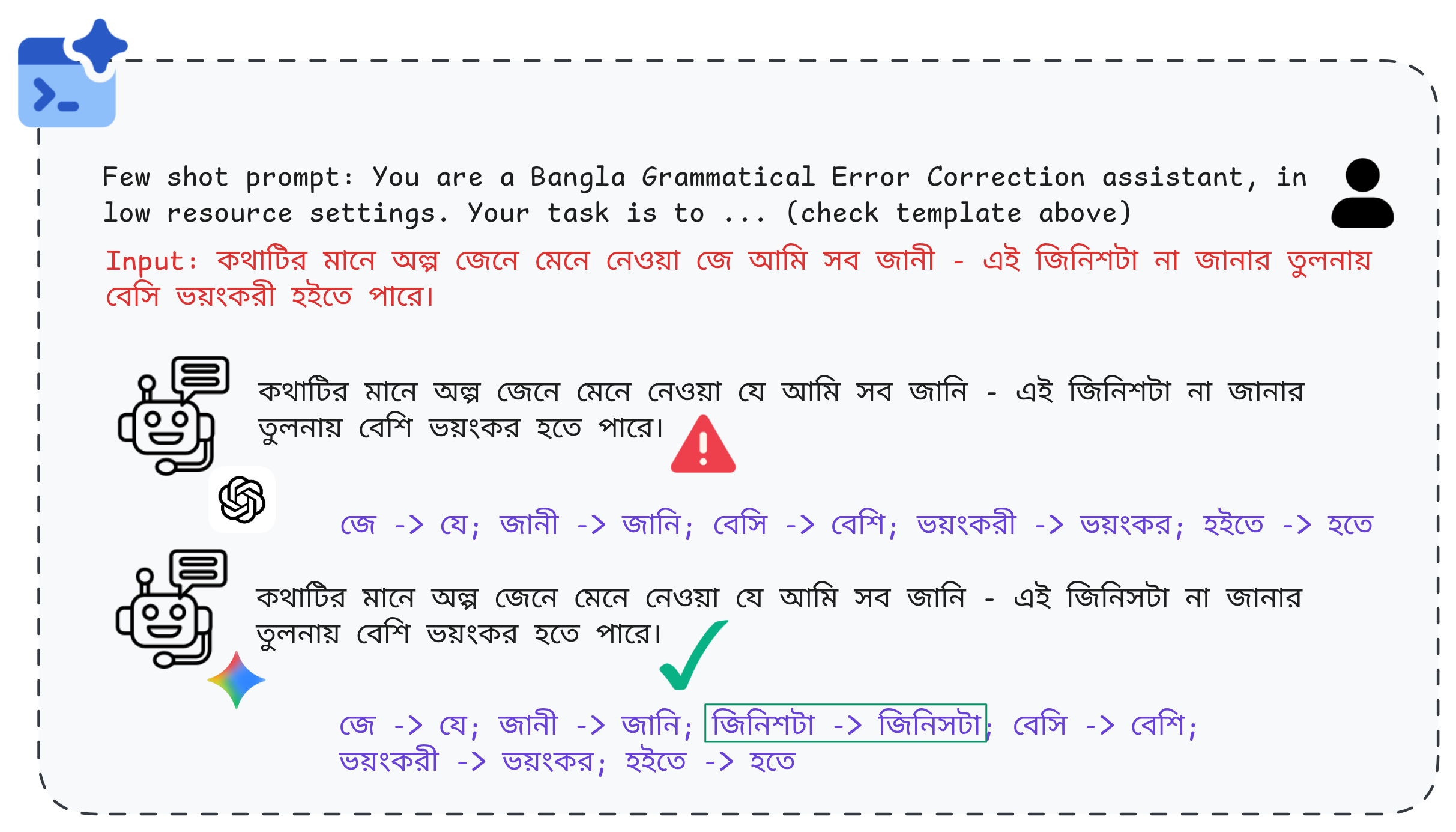}
  \caption {\label{A-1} Comparison of model outputs on a \textbf{multi-correction \textsc{BN} example} (Transl. Knowing only a little about something and assuming that I know everything can be more dangerous than not knowing at all.) from test set.  \textit{Gemini's output fully aligns with the gold standard, while GPT omits one necessary edit.}}
\end{figure*}

\begin{figure*}[ht]
    \centering
    \begin{minipage}[b]{0.48\linewidth}
        \centering
        \includegraphics[width=\linewidth]{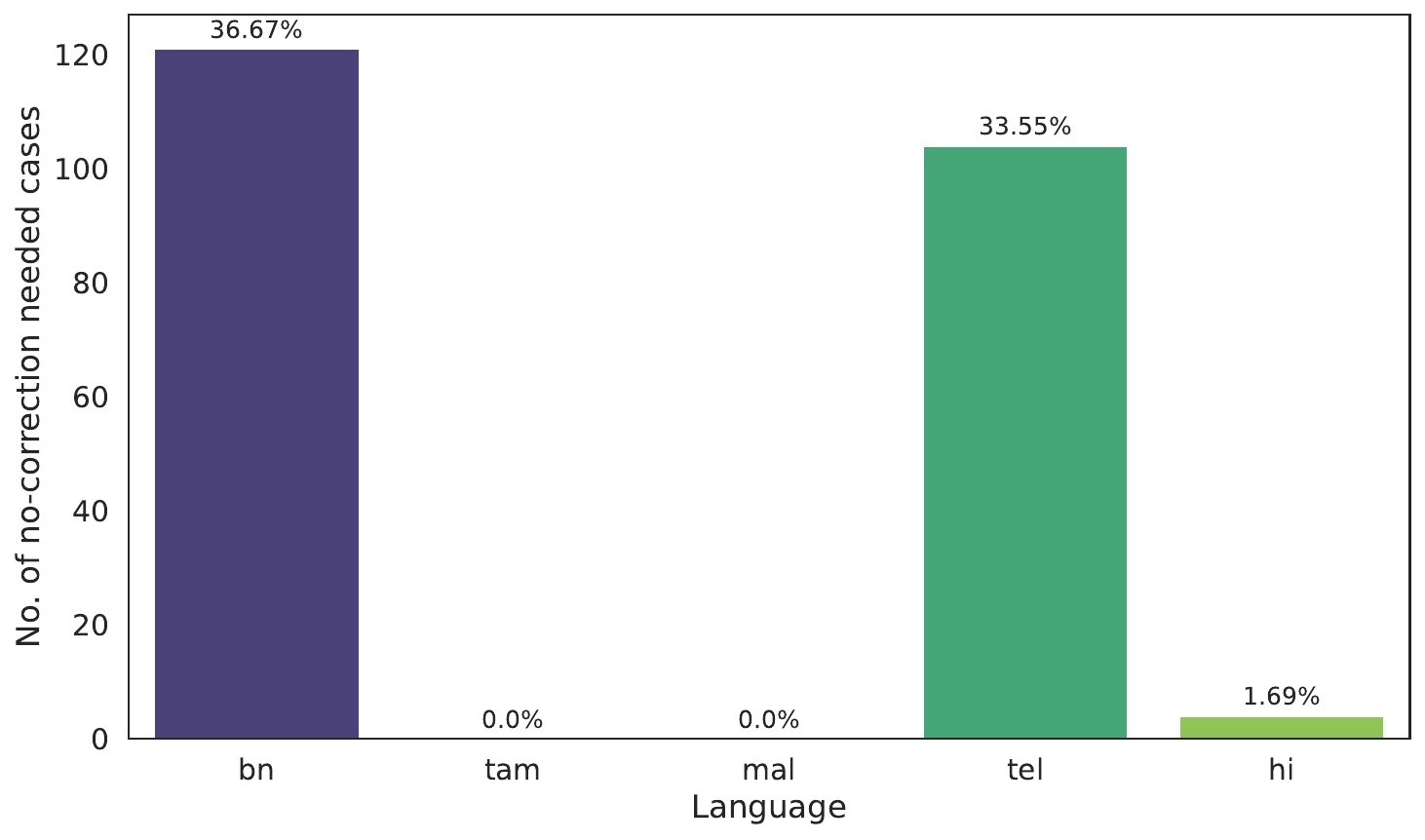}
    \end{minipage}
    \hfill
    \begin{minipage}[b]{0.48\linewidth}
        \centering
        \includegraphics[width=\linewidth]{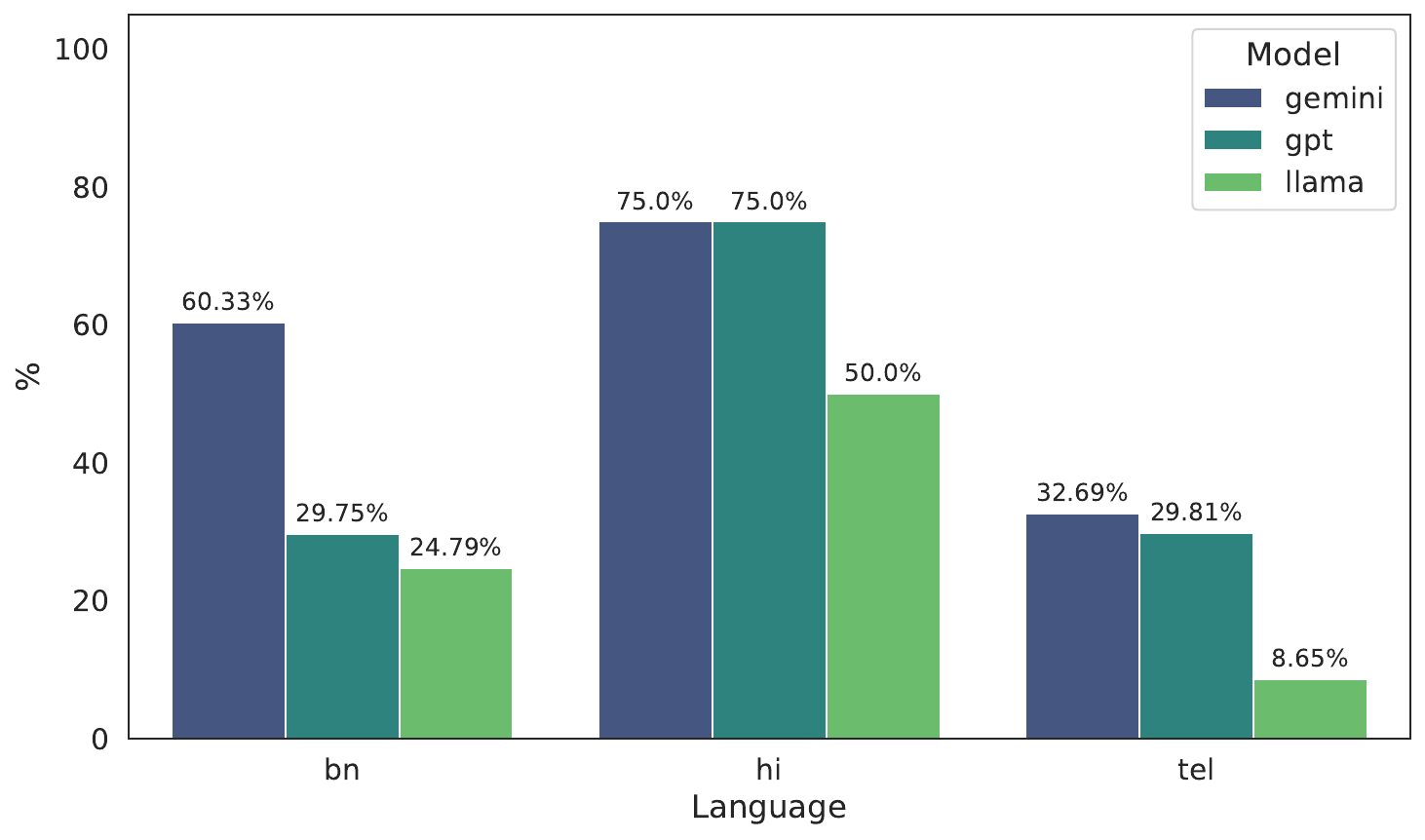}
    \end{minipage}
    \caption{\label{A-2} L: Distribution of test set cases where no corrections are needed. R: How well the models followed the instruction by not editing when no changes were required?}
\end{figure*}

\section{Tokenization in LLMs}
\label{sec:appendix2}
Tokenization--the process of mapping raw text to a sequence of numerical input IDs--is a primary determinant of computational efficiency in Large Language Models (LLMs). This is particularly critical for Indic languages, where complex morphology and non-Latin scripts often result in ``over-fragmentation'' by standard Byte-Pair Encoding (BPE) algorithms. A high Fertility Score (defined as the average number of tokens required to represent a single word) has two direct consequences for GEC: increased latency and Context Window Reduction.

\begin{figure*}[ht]
  \includegraphics[width=1.0\linewidth]{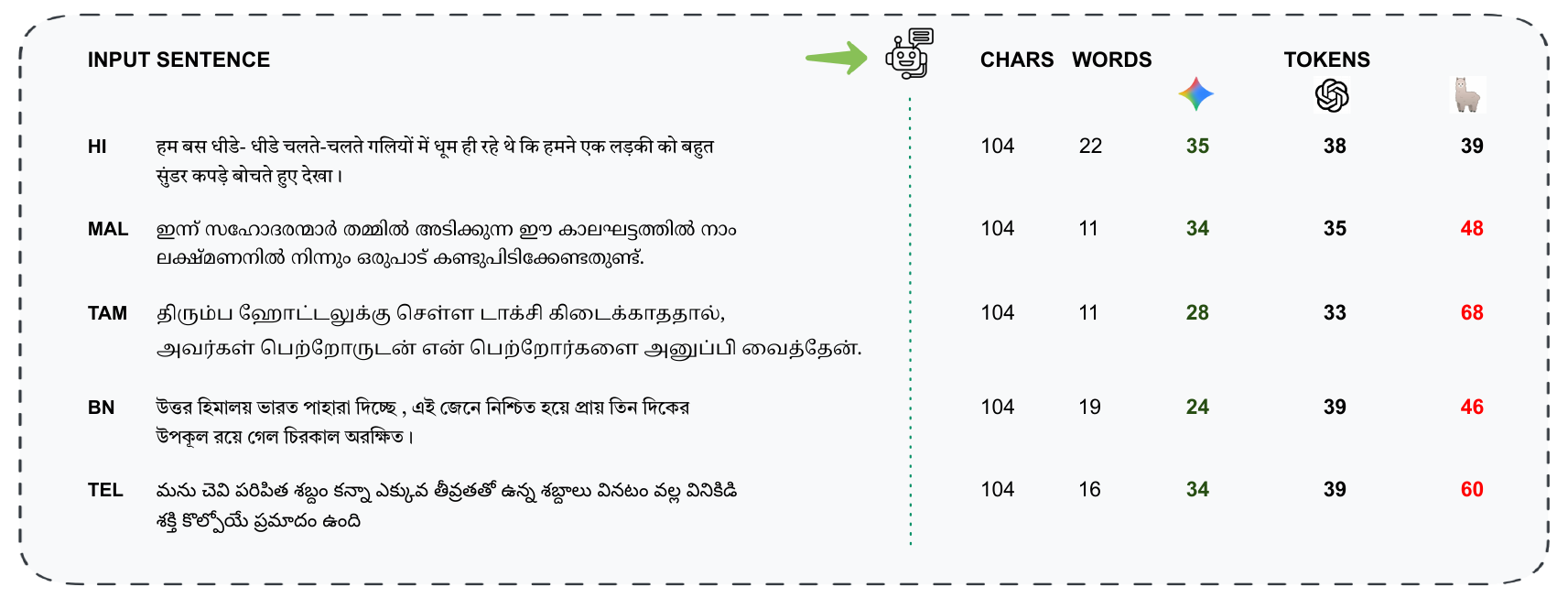}
  \caption {\label{A-3} Tokenization density across the three architectures}
\end{figure*}

To quantify this impact, we calculated the fertility score $F$ (token-to-word ratio) across our complete test sets for the five Indic languages. We compared the tokenizers of the three state-of-the-art models used in this study:
\begin{itemize}
    \item GPT-4.1 Mini: Uses the \verb|o200k\_base| tokenizer, an expanded vocabulary BPE designed to improve multilingual compression over the previous \verb|cl100k| standard.
    \item Gemini 2.5 Flash: Uses a \verb|SentencePiece|-based multimodal tokenizer optimized for high-throughput processing. It supports implicit caching, a feature that automatically identifies and reuses repeated parts of a prompt to reduce costs and latency. 
    \item Llama-4-Maverick: Uses the meta-llama/\verb|Llama-4-Maverick-17B-Instruct| tokenizer, representing the current state-of-the-art in open-weight models.
\end{itemize}
To provide a qualitative perspective on this variance, Figure \ref{A-3} visualizes the raw tokenization of five input sentences taken from the test set (one per language). To ensure a good comparison, we chose samples with input length of exactly 104 unicode codepoints (not visual graphemes, Indic scripts rely on complex rendering rules where a single visual character is often composed of multiple codepoints).

Table \ref{tab:fertility_comparison} presents the comparative fertility scores. Our analysis highlights a significant ``Tokenization Gap'' between proprietary and open-weight models for Dravidian languages.\\\\
\textbf{Key Observations}\\\\
1. The Gemini tokenizer consistently achieves the lowest fertility scores, demonstrating superior compression for Indic scripts. Notably, it achieves $F$ 1.76 for Bengali, providing a 24\% efficiency gain over GPT-4.1 Mini and a 36\% gain over Llama 4.\\
2. The Dravidian Disparity: While Indo-Aryan languages (Hindi, Bengali) are tokenized efficiently across all models, Dravidian languages suffer from significantly higher segmentation.\\
3. Llama 4 Performance Gap: The open-weight Llama 4 tokenizer exhibits severe over-fragmentation for Tamil ($F=5.88$) and Malayalam ($F=4.58$). Correcting a Tamil sentence using Llama 4 requires processing 2.3$\times$ more tokens than Gemini 2.5 Flash. This suggests that while Llama 4 is performant in reasoning, its tokenizer acts as a bottleneck for inference speed and cost in Dravidian GEC tasks, likely due to a higher reliance on byte-level fallback mechanisms for these scripts.

\begin{figure*}[ht]
  \includegraphics[width=1.0\linewidth]{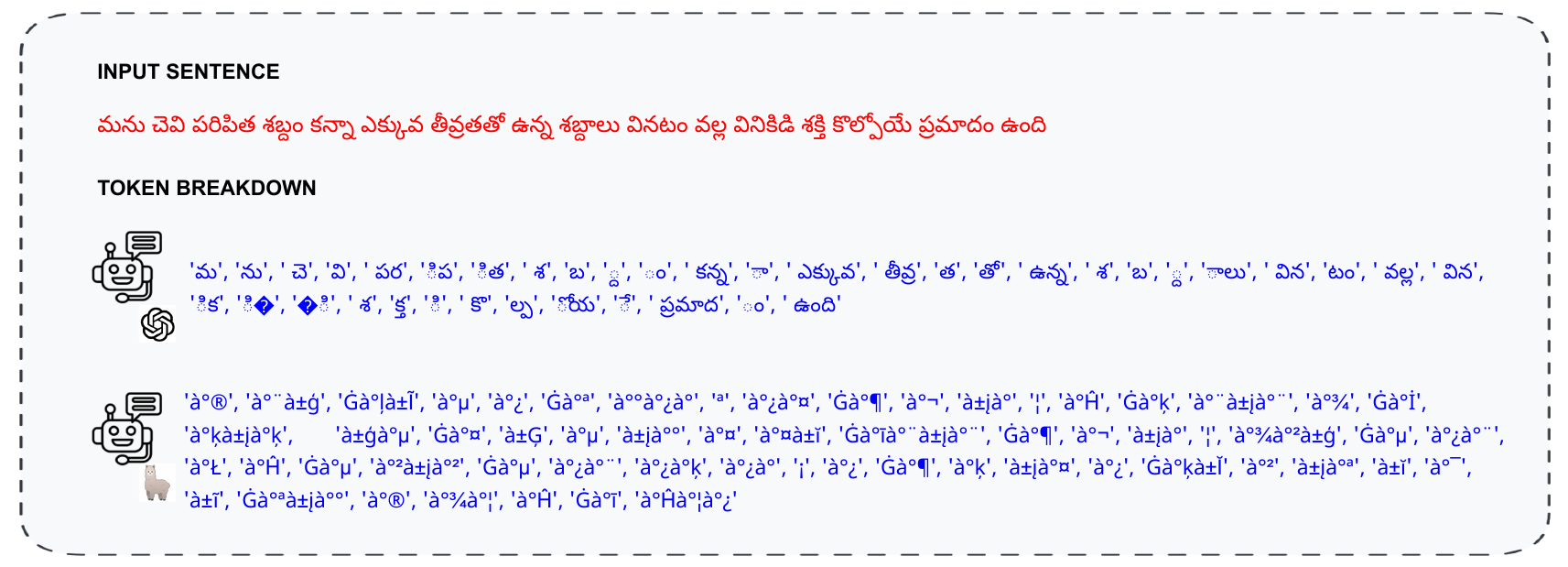}
  \caption {\label{A-4} Breakdown of tokenization in GPT-4.1-mini vs. Llama-4-Maverick, for the \textsc{TEL} input. (Gemini 2.5 Flash API does not currently expose token-level breakdown information.)}
\end{figure*}

\begin{table*}[ht]
\centering
\caption{\textbf{Cross-Model Tokenizer Fertility Comparison.} Lower scores imply better efficiency, lower latency and reduced inference cost.}
\label{tab:fertility_comparison}
\begin{tabular}{l c c c c}
\toprule
\rowcolor{black!5}
\textbf{Language} & \textbf{Script Family} & \textbf{GPT-4.1 Mini} & \textbf{Gemini 2.5 Flash} & \textbf{Llama 4 Maverick} \\
\midrule
\textsc{HI}      & Devanagari     & 1.44 & \textbf{1.31} & 1.55 \\
\textsc{BN}      & Eastern Nagari & 2.32 & \textbf{1.76} & 2.77 \\
\textsc{TAM}     & Dravidian      & 3.09 & \textbf{2.54} & \underline{\textcolor{red}{5.88}} \\
\textsc{TEL}     & Dravidian      & 2.97 & \textbf{2.87} & 4.32 \\
\textsc{MAL}     & Dravidian      & 3.20 & \textbf{3.10} & 4.58 \\
\bottomrule
\end{tabular}
\end{table*}
\end{document}